\documentclass{article}
\usepackage{ijcai21}

\usepackage{times}

\usepackage{soul}
\usepackage{url}
\usepackage[hidelinks]{hyperref}
\usepackage[utf8]{inputenc}
\usepackage[small]{caption}
\usepackage{graphicx}
\usepackage{amsmath}
\usepackage{booktabs}
\usepackage{xspace}
\usepackage[dvipsnames, table]{xcolor}
\usepackage{color}
\usepackage{amsfonts}
\usepackage{adjustbox}
\usepackage{subcaption}
\usepackage{paralist}
\usepackage{multirow}

\newcommand{\ie}{\emph{i.e.}}
\newcommand{\cp}{\textsc{CP}\xspace}
\newcommand{\csp}{\textsc{CSP}\xspace}
\newcommand{\cop}{\textsc{COP}\xspace}
\newcommand{\efsp}{\textsc{EFSP}\xspace}
\newcommand{\efop}{\textsc{EFOP}\xspace}

\newcommand{\ghost}{\textsc{GHOST}\xspace}
\newcommand{\cppn}{\textsc{CPPN}\xspace}
\newcommand{\icn}{\textsc{ICN}\xspace}

\newcommand{\cproblem}[3]%
{\begin{trivlist}
  \item[]%
    \textbf{Problem:} \textsc{#1}\\
    \textit{Input:} #2\\
    \textit{Question:} #3
  \end{trivlist}%
}

\begin{document}

\title{Learning Interpretable Error Functions\\for Combinatorial Optimization Problem Modeling}

\author{
Florian Richoux$^1$\footnote{Contact Author}\and
Jean-François Baffier$^2$\\
\affiliations
$^1$AIST, Japan / JFLI, CNRS, Japan\\
$^2$The University of Tokyo, Japan  / JFLI, CNRS, Japan\\
\emails
florian.richoux@aist.go.jp,
jf@baffier.fr
}


\maketitle

\begin{abstract}
  In Constraint  Programming, constraints  are usually  represented as
  predicates allowing  or forbidding combinations of  values. However,
  some algorithms exploit a finer representation: error functions.
  Their usage comes with a price though: it makes problem modeling
  significantly harder.
  Here, we propose  a method to automatically learn  an error function
  corresponding  to  a  constraint,   given  a  function  deciding  if
  assignments  are  valid  or  not.   This is,  to  the  best  of  our
  knowledge, the first attempt  to automatically learn error functions
  for hard constraints.  Our method
  uses   a  variant   of  neural   networks  we   named  Interpretable
  Compositional Networks,  allowing us  to get  interpretable results,
  unlike regular artificial neural networks.   
  Experiments  on 5  different constraints  show that  our system  can
  learn functions that scale to  high dimensions, and can learn fairly
  good functions over incomplete spaces.
\end{abstract}

\section{Introduction}\label{sec:introduction}


Twenty  years separate  Freuder's  papers \shortcite{Freuder1997}  and
\shortcite{Freuder2018},  both about  the grand  challenges Constraint
Programming (\cp) must tackle \emph{``to be pioneer of a new usability
  science and to go on to engineering usability''}~\cite{Freuder2007}.
To  respond  to   the  lack  of  a  ``Model  and   Run''  approach  in
\cp~\cite{Puget2004,Wallace2003},   several    languages   have   been
developed  since  the  late 2000's,  such  as  ESSENCE~\cite{essence},
XCSP~\cite{XCSP-paper}  or  MiniZinc~\cite{minizinc}.   However,  they
require users to have deep expertise on global constraints and to know
how well  these constraints, and  their associated mechanisms  such as
propagators,  are suiting  the  solver.   We are  still  far from  the
original Holy Grail  of \cp: \emph{``the user states  the problem, the
  computer solves it''}~\cite{Freuder1997}.

This paper makes a contribution  in automatic \cp problem modeling. We
focus  on Error  Function  Satisfaction and  Optimization Problems  we
defined  in  the  next   section.   Compare  to  classical  Constraint
Satisfaction  and Constrained  Optimization Problems,  they rely  on a
finer structure about  the problem: the cost  functions network, which
is, for this work, an  ordered structure over invalid assignments that
constraint solver can exploit efficiently to improve the search.


In  this  paper,  we  propose   a  method  to  learn  error  functions
automatically; a direction that, to the best of our knowledge, had not
been   explored  in   Constraint  Programming.    We  focus   here  on
``easy-to-use'' aspects of Constraint Programming.

To sum up, the  main contributions of this paper are:  1.  to give the
first formal  definition of  Error Function Satisfaction  Problems and
Error Function  Optimization Problems, 2.  to  introduce Interpretable
Compositional  Networks,   a  variant   of  neural  networks   to  get
interpretable results, 3.  to propose an architecture of Interpretable
Compositional Network to  learn error functions, and 4.   to provide a
proof  of  concept  by learning  Interpretable  Compositional  Network
models of error functions, using a genetic algorithm, and to show that
most of  models give scalable  functions, and remain  fairly effective
using incomplete training sets.

\section{Error Function and Optimization Problems}\label{sec:efsp}

Constraint  Satisfaction Problem  (\csp) and  Constrained Optimization
Problem  (\cop)  are hard  constraint-based  problems  defined upon  a
classical  constraint  network,  where  constraints  can  be  seen  as
predicates  allowing  or  forbidding  some  combinations  of  variable
assignments.  

Likewise,  Error  Function  Satisfaction  Problem  (\efsp)  and  Error
Function  Optimization  Problem   (\efop)  are  hard  constraint-based
problems  defined  upon  a  specific  constraint  network  named  cost
function network~\cite{GuidedTour}.
Constraints    are     then    represented    by     cost    functions
\(f: D_1  \times D_2 \times  \ldots \times D_n \rightarrow  E\), where
\(D_i\) is  the domain of  \(i\)-th variable in the  constraint scope,
\(n\) the number of variables (\ie,  the size of this scope) and \(E\)
the set of possible costs.

A cost function network is a quadruplet \(\langle V, D, F, S \rangle\)
where \(V\) is a  set of variables, \(D\) the set  of domains for each
variable, \ie,  the sets of values  each variable can take,  \(F\) the
set of cost functions and \(S\)  a cost structure. A cost structure is
also a quadruplet \(S = \langle  E, \oplus, \bot, \top \rangle\) with
\(E\)  the  totally ordered  set  of possible  costs, \(\oplus\)  a
commutative,  associative,  and   monotone  aggregation  operator  and
\(\bot\)  and  \(\top\) the  neutral  and absorbing  elements  of
\(\oplus\), respectively.

In Constraint Programming, cost functions are often associated to soft
constraints:  they can  be interpreted  as preferences  over valid  or
acceptable assignments.  However, this is not necessarily the case: it
depends  on the  cost  structure.  For  instance,  the classical  cost
structure
\[S_{t/f} = \langle \{true, false\}, \wedge, true, false\rangle\] makes
the  cost  function  network  equivalent  to  a  classical  constraint
network, so dealing with hard constraints.

Here,  we  consider  particular  cost functions  that  represent  hard
constraints  only,   by  considering   the  additive   cost  structure
\(S_+ = \langle \mathbb{R}, +,  0, \infty\rangle\).  The additive cost
structure produces  useful cost  function networks  capturing problems
such as Maximum Probability Explanation (MPE) in Bayesian networks and
Maximum    A   Posteriori    (MAP)   problems    in   Markov    random
fields~\cite{Hurley16}.

In this paper, an {\bf error function} is a cost function defined in a
cost  function  network  with  the  additive  cost  structure~\(S_+\).
Intuitively,  error  functions  are  preferences  over  \emph{invalid}
assignments.   Let  \(f_c\)  be   an  error  function  representing  a
constraint \(c\)  and \(\vec x_c\) be an assignment of variables  in the
scope of \(c\).  Then \(f_c(\vec x_c) = 0\) iff \(\vec x_c\) satisfies
the  constraint  \(c\).   For all  invalid  assignments~\(\vec  i_c\),
\(f_c(\vec i_c) > 0\) such that  the closer \(f_c(\vec i_c)\) is to 0,
the closer~\(\vec i_c\) is to satisfy \(c\).

The goal of  this paper is {\bf  not} to study the  advantages of such
cost  function networks  over  regular  constraint networks.   Without
formally defining  \efsp and  \efop problems, some  studies illustrate
that  solvers   (in  particular,  metaheuristics)  can   exploit  this
structure   efficiently  leading   to  state-of-the-art   experimental
results,     both     in     sequential~\cite{AS}     and     parallel
solving~\cite{caniou2015constraints}.   In addition,  our Experiment~3
shows  that  error  functions representing  the  classic  AllDifferent
constraint gives models  that clearly outperformed a model  based on a
regular  constraint networks  in terms  of runtimes,  for models  with
either hand-crafted or learned error functions.

Let \(\vec x\) be a variable assignment,
and denote by \(\vec x_c\) the projection of \(\vec x\) over variables
in the scope of  a constraint \(c\).  We can now  define the \efsp and
\efop problems.

\cproblem%
{Error Function Satisfaction Problem}%
{ A cost function network \(\langle V, D, F, S_+ \rangle\).}%
{ Does   a   variable   assignment   \(\vec   x\)   exist   such   that
  \(\forall f_c \in F,\ f_c(\vec x_c)=0\) holds?}%

\cproblem%
{Error Function Optimization Problem}%
{ A  cost function  network \(\langle  V, D,  F, S_+  \rangle\) and  an
  objective function \(o\).}%
{ Find a  variable assignment \(\vec  x\) maximizing or  minimizing the
  value        of        \(o(\vec         x)\)        such        that
  \(\forall f_c \in F,\ f_c(\vec x_c)=0\) holds.}%

With the  system we  propose in  this paper,  users provide  the usual
constraint network  \(\langle V, D,  C \rangle\), and it  computes the
equivalent cost  function networks \(\langle  V, D, F,  S_+ \rangle\).
Learned error functions composing the set \(F\) are independent of the
number  of variables  in constraints  scope, and  are expressed  in an
interpretable  way: users  can understand  these functions  and easily
modify them at will.  This way, users  can have the power of \efsp and
\efop with the same modeling effort as for \csp and \cop.


\section{Related works}\label{sec:related}

This work belongs to one of the three directions identified by Freuder
\shortcite{Freuder2007}:    \emph{Automation},    \ie,    ``automating
efficient and  effective modeling and  solving''.  To the best  of our
knowledge, few efforts have been done on the modeling side.

Another  of  these  three  directions which  is  slightly  related  is
\emph{Acquisition} described  by Freuder to be  ``acquiring a complete
and correct representation of  real problems''.  Remarkable efforts on
this topic  have been done  by Bessiere's research team,  for instance
with  constraints learning  by  induction from  positive and  negative
examples~\cite{Bessiere05}  and  with  interactive  queries  asked  to
users~\cite{Bessiere07},  and with  constraint  network learning  also
through  with  interactive queries~\cite{Bessiere13}.

Model Seeker~\cite{Beldiceanu12}  is a passive learning  system taking
positive  examples  only, which  are  certainly  easier for  users  to
provide.   It transforms  examples  into data  adapted  to the  Global
Constraint   Catalog~\cite{catalog},   then   generate   and   simplify
candidates by eliminating dominated ones. Model Seeker is particularly
efficient  to find  a good  inner structure  of the  target constraint
network.

Teso~\shortcite{Teso19}  gives a  good survey  on Constraint  Learning
with   this    interesting   remark:    ``A   major    bottleneck   of
[constraint-based  problem  modeling]  is   that  obtaining  a  formal
constraint theory  is non-obvious:  designing an  appropriate, working
constraint satisfaction or optimization problem requires both domain
and  modeling expertise.  For this  reason, in  many cases  a modeling
expert is  hired and  has to  interact with  domain expert  to acquire
informal requirements  and turn them  into a valid  constraint theory.
This process can be expensive and time-consuming.''

We can  consider that Constraint Acquisition,  or Constraint Learning,
focuses on modeling expertise and puts domain expertise on background:
users  would not  be able  to understand  and modify  a learned  model
without the  help of a modeling  expert. The goal of  these systems is
mainly to simplify the interaction between the domain and the modeling
experts.

Our  work  is  taking  the  opposite direction:  we  focus  on  domain
expertise and put modeling expertise on background. 
With our system, users always have the control
over constraints' representation, which can be modified at will to fit
needs   related   to   their   domain   expertise.    \emph{Constraint
  Implementation Learning} is what best describes this research topic.

\section{Method design}\label{sec:method}

The main result of this paper  is to propose a method to automatically
learn an error function representing  a constraint, to make easier the
modeling of \efsp/\efop.   We are tackling a  regression problem since
the goal  is to find a  function that outputs a  target value.  Before
diving into the  description of our method, we need  to introduce some
essential notions.


\subsection{Definitions}


We propose a method to automatically  learn an error function from the
\emph{concept}  of   a  constraint.   As  described   in  Bessiere  et
al.~\shortcite{Bessiere2017}, the \textbf{concept}  of a constraint is
a  Boolean function  that,  given an  assignment  \(\vec x\),  outputs
\emph{true} if  \(\vec x\) satisfies the  constraint, and \emph{false}
otherwise. Concepts  are the  predicate representation  of constraints
referred at the beginning of Section~\ref{sec:efsp}.  


Our method learns  error functions in a  supervised fashion, searching
for   a   function  computing   the   \emph{Hamming   cost}  of   each
assignment. The \textbf{Hamming cost} of  an assignment \(\vec x\) is the
minimum  number  of  variables  in  \(\vec x\)  to  reassign  to  get  a
\textbf{solution}, \ie, a variable assignment
satisfying  the  considered constraint.  
If \(\vec x\) is a solution, then  its Hamming cost is 0.  

Given the  number of variables of  a constraint and their  domain, the
\textbf{constraint   assignment  space}   is   the   set  of   couples
\((\vec  x, b)\)  where  \(\vec x\)  is an  assignment  and \(b\)  the
Boolean output of the concept  applied on \(\vec x\).  Such constraint
assignment spaces  can be  generated from  concepts. These  spaces are
said to be \textbf{complete} if and  only if they contain all possible
assignments, \ie, all combinations of  possible values of variables in
the  scope  of the  constraint.   Otherwise,  spaces  are said  to  be
\textbf{incomplete}.

In  this work,  we  consider an  error function  to  be a  (non-linear)
combination of elementary  operations. 
Complete spaces are intuitively good training sets since it is easy to
compute the  exact Hamming cost  of their elements.  We  also consider
assignments from incomplete  spaces where their Hamming  cost has been
approximated  regarding  a  subset  of  solutions  in  the  constraint
assignment space, in case the exact Hamming cost function is unknown.


\subsection{Main result}\label{sec:model}

To learn an  error function as a non-linear  combination of elementary
operations,   we  propose   a   network   inspired  by   Compositional
Pattern-Producing   Networks   (\cppn).   \(\cppn\)s~\cite{CPPN}   are
themselves a variant of artificial  neural networks.  While neurons in
regular neural  networks usually  contain sigmoid-like  functions only
(such  as ReLU,  \ie{}  Rectified Linear  Unit),  \cppn's neurons  can
contain   many  other   kinds   of   function:  sigmoids,   Gaussians,
trigonometric   functions,   and   linear  functions   among   others.
\(\cppn\)s are often used to generate  2D or 3D images by applying the
function modeled by  a \cppn giving each pixel  individually as input,
instead of considering  all pixels at once.  This  simple trick allows
the learned \cppn model to produce images of any resolution.

We  propose our  variant by  taking these  two principles  from \cppn:
having neurons containing one operation  among many possible ones, and
handling  inputs   in  a  size-independent  fashion.    Due  to  their
interpretable  nature,  we  named  our  variant  \textbf{Interpretable
  Compositional Networks}  (\icn).  In  this paper, our  \(\icn\)s are
composed of  four layers, each of  them having a specific  purpose and
themselves composed of neurons applying  a unique operation each.  All
neurons  from  a  layer  are  linked to  all  neurons  from  the  next
layer. The weight on each link is purely binary: its value is either 0
or 1.  This restriction is  crucial to obtain interpretable functions.
A weight  between neurons \(n_1\) and  \(n_2\) with the value  1 means
that the neuron  \(n_2\) from layer \(l+1\) takes as  input the output
of the neuron \(n_1\) from layer  \(l\). Weight with the value 0 means
that \(n_2\) discards the output of \(n_1\).

Here is our method workflow in 4 points:

1. Users  provide a  regular constraint
network \(\langle V,  D, C \rangle\) where \(C\) is  a set of concepts
representing constraints.

2. For each constraint concept \(c\), we generate its \icn input space
\(X\), which is either a  complete or incomplete constraint assignment
space.  Those  input spaces  are our  training sets.  If the  space is
complete, then the Hamming cost of each assignment can be pre-computed
before learning  our \icn  model. Otherwise,  the incomplete  space is
composed of  randomly drawn assignments  and only an  approximation of
their Hamming cost can be pre-computed.

3.  We  learn  the  weights of  our  \icn  model  in  a
supervised fashion, with the following loss function:
\begin{equation}\label{eq:loss}
\text{loss} = \sum_{\vec x \in X} \left(|ICN(\vec x) - Hamming(\vec x)|\right) + R(ICN)
\end{equation}

where \(X\) is  the constraint assignment space,  \icn(\(\vec x\)) the
output  of  the \icn  model  giving  \(\vec x  \in  X\)  as an  input,
Hamming(\(\vec x\)) the pre-computed Hamming  cost of \(\vec x\) (only
approximated if \(X\) is incomplete),  and R(\icn) is a regularization
between  0  and  0.9  to  favor short  \(\icn\)s,  \ie,  with  as  few
elementary operations as possible,  such that \(R(ICN) = 0.9
\times \frac{\text{Number of selected elementary operations}}{\text{Maximal
    number of elementary operations}}\).

4.  We have  hold-out test
sets of assignments from larger  dimensions to evaluate the quality of
our learned error functions.

Notice we also have a hold-out validation set to fix the values of our
hyperparameters, as described in Section~\ref{sec:ga}.

\begin{figure}
  \centering
  \includegraphics[width=\linewidth]{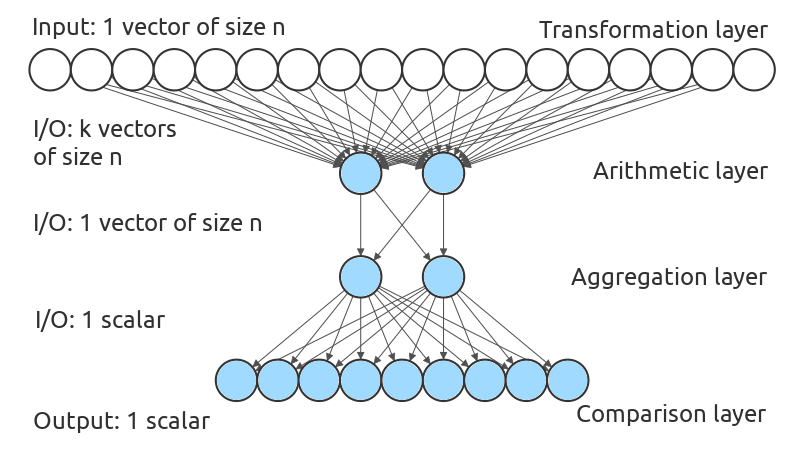}
  \caption{Our 4-layer network. Layers with blue neurons have mutually exclusive operations.}
  \label{fig:model}
\end{figure}

Figure~\ref{fig:model}   is   a   schematic  representation   of   our
network. It  takes as input an  assignment of \(n\) variables,  \ie, a
vector    of    \(n\)    integers.    The    first    layer,    called
\textbf{transformation  layer},  is   composed  of  18  transformation
operations, each of  them applied element-wise on each  element of the
input vector.   Such operations are  for instance the  maximum between
the  \(i\)-th and  \(i+1\)-th elements  of  the input  vector, or  the
number of  \(j\)-th elements of  the vector smaller than  the \(i\)-th
element such  that \(j >  i\) holds.  This  layer is composed  of both
linear and non-linear  operations.  If an operation  is selected (\ie,
it has an outgoing  weight equals to 1), it outputs  a vector of \(n\)
integers.


If \(k\)  transformation operations  are selected,  then the  next layer
gets  \(k\)  vectors of  \(n\)  integers  as  input.   This layer  is  the
\textbf{arithmetic layer}.  Its  goal is to apply  a simple arithmetic
operation in a component-wise fashion on all \(i\)-th element of our \(k\)
vectors  to get  one  vector of  \(n\) integers  at  the end,  combining
previous transformations into a unique vector. We have considered only
2 arithmetic operations so far: the addition and the multiplication.


The output of the arithmetic layer is given to the \textbf{aggregation
  layer}. This layer crunches the  whole vector into a unique integer.
At the  moment, the  aggregation layer is  composed of  2 operations:
\emph{Sum} computing  the sum of input  values and \emph{\(Count_{>0}\)}
counting the number of input values strictly greater than 0.


Finally, the computed scalar  is transmitted to the \textbf{comparison
  layer}  with 9  operations.  Examples  of these  operations are  the
identity,  or  the   absolute  value  of  the  input   minus  a  given
parameter. This  layer compares its  input with an  external parameter
value, or the number of variables  of the problem, or the domain size,
among others.


All elementary operations  in our model are generic: we  do not choose
them to  fit one or  several particular  constraints. Due to  the page
limit, we  cannot give a  comprehensive list of the  18 transformation
and 9 comparison  operations here.  Although an in-depth  study of the
elementary operations properties would be  interesting, this is out of
the  scope  of  this  paper:  its   goal  is  to  show  that  learning
interpretable error functions  via a generic \icn is  possible, and in
the same way results with neural networks do not always use ReLU as an
activation function, there is no reason  to reduce \icn to its current
31  elementary  operations  or  even  a  4-layer  architecture.   Such
elements can be changed by users to best fit their needs.

To  have  simple   models  of  error  functions,   operations  of  the
arithmetic, the  aggregation, and  the comparison layers  are mutually
exclusive, meaning that  precisely one operation is  selected for each
of  these layers.   However, many  operations from  the transformation
layer can be selected to compose the error function. Combined with the
choice  of  having  binary  weights,  it allows  us  to  have  a  very
comprehensible combination of elementary  operations to model an error
function,
making it readable  and intelligible by a human  being.  For instance,
the     most      frequenly     learned     error      function     is
\(Count_{>0}\big(|\{j\  |\ x[j]=x[i]  \text{ and  } j>i\}|\big)\)  for
AllDifferent,  and  \(Euclidian_p\big(\sum^{n}_{i=1}  x[i]\big)\)  for
LinearSum,       \ie,       the      Euclidian       division       of
\(\big(\sum^{n}_{i=1} x[i] - p\big)\) by the maximal domain size, with
a parameter \(p\) equals to the right hand side constant of the linear
equation. Thus, once the model of  an error function is learned, users
have  the choice  to  run the  network in  a  feed-forward fashion  to
compute  the error  function,  or  to re-implement  it  directly in  a
programming  language.   Users  can  use  our  system  to  find  error
functions  automatically, but  they  can  also use  it  as a  decision
support system to find promising  error functions that they may modify
and adapt by hand.

\subsection{Learning with Genetic Algorithms}\label{sec:ga}

Like any neural network, learning an error function through an
\icn 
boils  down  to learning  the  value  of  its  weights.  Many  of  our
elementary    operations    are    discrete,   therefore    are    not
differentiable. Then,  we cannot  use a back-propagation  algorithm to
learn the \icn's  weights. This is why we use  a genetic algorithm for
this task.


Since our weights are binary,  we represent individuals of our genetic
algorithm by a binary vector, 
each bit corresponding to one  operation in the four layers indicating
if the operation is selected to be part of the error
function.  

We randomly generate  an initial population of  160 individuals, check
and fix them if they do  not satisfy the mutually exclusive constraint
of the comparison layer. Then, we run the genetic algorithm to produce
at  most  800  generations   before  outputting  its  best  individual
according to our fitness function.

Our genetic algorithm is  rather simple: The \textbf{fitness function}
is  the  loss   function  of  our  supervised   learning  depicted  by
Equation~\ref{eq:loss}.   \textbf{Selection} is  made by  a tournament
selection  between 2  individuals.   \textbf{Variation} is  done by  a
one-point crossover operation and  a one-flip mutation operation, both
crafted  to  always produce  new  individuals  verifying the  mutually
exclusive constraint of  the comparison layer.  The  crossover rate is
fixed  at 0.4,  and  exactly  one bit  is  mutated  for each  selected
individual with a mutation rate  of 1. \textbf{Replacement} is done by
an elitist  merge, keeping 17\% of  the best individuals from  the old
generation into the new one,  and a deterministic tournament truncates
the new  population to  160 individuals.   The algorithm  stops before
reaching 800 generations if no improvements have been done in the last
50    generations.    We    use   the    framework    \textsc{Evolving
  Objects}~\cite{EO} to code our genetic algorithm.

Our hyperparameters, \ie,  the population size, the  maximal number of
generations, the number  of steady generations before  early stop, the
crossover, mutation and replacement rates, and the size of tournaments
have been  chosen using ParamILS~\cite{paramils}, trained  one week on
one CPU over  a large range of values for  each hyperparameter. We use
the  same training  instance used  for Experiment  1, and  new, larger
instances as a hold-out validation
set. 
These  instances have  been chosen  because they  are larger  than our
training instances  and each  of them  contains about  4\(\sim\)5\% of
solutions,  which is  significantly  less than  the 10\(\sim\)20\%  of
solutions in training instances.

\section{Experiments}\label{sec:xp}
To  show the  versatility of  our method,  we tested  it on  five very
different constraints: AllDifferent,  Ordered, LinearSum, NoOverlap1D,
and Minimum.  According  to XCSP specifications~\cite{XCSP-paper}\footnote{see
also  \url{http://xcsp.org/specifications}}, those  global constraints
belong  to  four  different  families:  Comparison  (AllDifferent  and
Ordered),     Counting/Summing     (LinearSum),     Packing/Scheduling
(NoOverlap1D)  and  Connection  (Minimum).  Again  according  to  XCSP
specifications,  these  five constraints  are  among  the twenty  most
popular and common constraints.  We  give a brief description of those
five constraints below:

\begin{itemize}
  \item \textbf{AllDifferent} ensures  that variables must all  be assigned
  to different values.
  \item \textbf{LinearSum}       ensures       that      the       equation
  \(x_1 + x_2 +  \ldots + x_n = p\) holds, with the  parameter \(p\) a given
  integer.
  \item \textbf{Minimum} ensures  that the  minimum value of  an assignment
  verifies  a given  numerical condition.  In this  paper, we  choose to
  consider that  the minimum value must  be greater than or  equals to a
  given parameter \(p\).
  \item \textbf{NoOverlap1D}  is considering  variables as  tasks, starting
  from a  certain time (their  value) and each  with a given  length \(p\)
  (their  parameter).   The  constraint   ensures  that  no   tasks  are
  overlapping,  \ie, for  all indexes  \(i,j  \in \{1,n\}\)  with \(n\)  the
  number   of   variables,  we   have   \(x_i   +   p_i  \leq   x_j\)   or
  \(x_j + p_j  \leq x_i\). To have  a simpler code, we  have considered in
  our system that all tasks have the same length \(p\).
\item  \textbf{Ordered} ensures  that an  assignment of  \(n\) variables
  \((x_1\mathrel{,} \ldots\mathrel{,}  x_n)\) must  be ordered,  given a
  total  order.  In  this paper,  we  choose the  total order  \(\leq\).
  Thus,  for  all   indexes  \(i,j  \in  \{1,n\}\),  \(i   <  j\)  implies
  \(x_i \leq x_j\).
\end{itemize}

\subsection{Experimental protocols}\label{sec:xp_proto}

We conducted three experiments, with  two of them requiring samplings.
These samplings have been done  using Latin hypercube sampling to have
a good diversity among drawn
assignments. 
When we  need to
sample  the same  number \(k\)  solutions and  non-solutions, we  draw
assignments until we  get \(k\) of solutions  and \(k\) non-solutions.

Due to  stochastic learning, all  learning and testing have  been done
100 times.  We did not re-run  batches of experiments to keep the ones
with  the best  results, as  it should  always be  the case  with such
experimental protocols.

All experiments have been  done on a computer with a  Core i9 9900 CPU
and 32 GB of RAM, running on Ubuntu 20.04. Programs have been compiled
with GCC with  the 03 optimization option. Our entire  system, its C++
source code, experimental setups, and the results files are accessible
on
GitHub\footnote{\url{https://github.com/richoux/LearningErrorFunctions/tree/1.1}
}.

\subsubsection{Experiment 1: scaling}

The goal  of this experiment is  to show that learned  error functions
scale to  high-dimensional constraints, indicating that  learned error
functions are independent of the size of the constraint scope.

For  this  experiment,  error  functions are  learned  upon  a  small,
complete constraint assignment space, composed of about 500\(\sim\)600
assignments and containing about 10\(\sim\)20\% of solutions.
For  each  constraint,  we  run   100  error  function  learning  over
pre-computed complete  constraint assignment space.  Then,  we compute
the test  error of these learned  error functions over a  sampled test
set.   Sampled   test  sets   contain  10,000  solutions   and  10,000
non-solutions, with 100 variables on domains of size 100, belonging to
a constraint  assignment space of size  \(100^{100}~=~10^{200}\), thus
greatly   larger  than   training  spaces   containing  500\(\sim\)600
assignments.  

We show  normalized mean training  and test errors: first,  we compute
the mean  error among  all assignments composing  the training  or the
test set. Then, we divide it  by the number of variables composing the
assignments. Indeed, having a mean error  of 5 on assignments with 100
variables and  10 variables is  significantly different: the  first one
indicates a mean error every 20 variables, the second a mean error one
in two variables.

\subsubsection{Experiment 2: learning over incomplete spaces}

If, for any reasons, it is not possible to build a complete constraint
assignment  space, a  robust system  must be  able to  learn effective
error functions upon large, incomplete  spaces where the exact Hamming
cost of their assignments is unknown.

In this experiment,  we built pre-sampled training  spaces by sampling
10,000  solutions   and  10,000  non-solutions  on   large  constraint
assignment  spaces of  size between  \(10^{12}\) and  \(10^{13}\), and
with  solution  rates from  \(0.15\)\%  to  \(2.10^{-7}\)\%. Then,  we
approximate the Hamming  cost of each non-solution  by computing their
Hamming distance with the closest  solution among the 10,000 ones, and
learn error functions on these  20,000 assignments and their estimated
Hamming  cost.  Like  for  Experiment~1, we  run  100 error  functions
learning of these pre-sampled incomplete spaces, so that each learning
relies on  the same  training set.  Finally,  we evaluate  the learned
error functions over the same test sets than Experiment~1.

\subsubsection{Experiment 3: learned error functions to solve problems}\label{sec:xp_proto3}


The goal of  this experiment is to assess that  learned error function
can effectively  be used to solve  toy problems. Here, we  use a local
search solver to solve Sudoku.

Sudoku is a puzzle  game where all numbers in the  same row, the same
column and  the same sub-square  must be  different. There, it  can be
modeled as  a satisfaction  problem using the  AllDifferent constraint
only. We run 100 resolutions of random \(9\times9\) and \(16\times16\)
Sudoku grids, with a timeout of  10 seconds. If no solutions have been
found within 10 seconds, we consider the run to be unsolved.

We  consider  the  mean  and  median  run-time  to  compare  different
representations  of   the  AllDifferent   constraint.   We   have  two
baselines: 1., a pure \csp model where constraints are predicates, and
2.,  an  \efsp  model  with an  efficient  hand-crafted  error  function
representing  AllDifferent.  We  compare those  with two  models using
error functions learned with our system to represent AllDifferent: a.,
our \efsp model  using the most frequently learned  error function from
the  previous experiments  and run  through  our neural  network in  a
fast-forward  fashion, and  b., our  \efsp  model with  the same  error
function  but  directly  hard-coded  in   C++.   The  solver  and  its
parameters remain the  same: the only thing that is  modified in these
four  different   models  is   the  expression  of   the  AllDifferent
constraint.

\subsection{Results}\label{sec:results}


\subsubsection{Experiments 1 \& 2}\label{sec:results_xp1}

Table~\ref{tab:training}  shows the  training errors  of Experiment~1,
where error functions have been learned 100 times for each constraint.
The first  column contains the  normalized mean training error  of the
most frequently  learned error function  among the 100 runs,  with its
frequency in  parenthesis. Next columns  concern the median,  the mean
and the standard deviation.

Table~\ref{tab:test} contains the normalized mean test errors of error
functions learned  with Experiments~1 and~2, with  their median, mean
and standard  deviation. The  normalized mean test  error of  the most
frequently  learned  error  function   for  each  constraint  in  each
experiment  has  been isolated  in  the  first  column of  number,  for
comparison.

Comparing   Table~\ref{tab:training}    and   the   first    half   of
Table~\ref{tab:test} lead  us to conclude  that our system is  able to
learn  error   functions  that  scale  for   most  constraint,  namely
AllDifferent, LinearSum  and Minimum. Their median  training errors in
Table~\ref{tab:training} are  perfect of  almost perfect, so  as their
median test errors on greatly larger constraint assignment spaces.

Results are not as good for  NoOverlap1D and Ordered, which are clearly
the  most  intrinsically  combinatorial  constraints  among  our  five
ones. One could  think that our system is overfitting  on its training
set, but results from Experiment~2 lead us to another conclusion.

To see this,  let's observe these Experiment~2's  results by comparing
the first and the second half of Table~\ref{tab:test}. Error functions
learned  over incomplete  training  spaces  are as  good  as the  ones
learned  over small  complete spaces  for LinearSum  and Minimum.   We
observe  significant improvements  of the  median and  the mean  for
NoOverlap1D   (36.07\%  and   55.76\%)   and   Ordered  (52.83\%   and
49.05\%). This  is due not  because error functions  from Experiment~1
were overfitting, but because spaces  from Experiment~1 were too small
for  these  highly  combinatorial   constraints,  containing  too  few
different combinations and Hamming cost patterns.

\begin{table}
  \centering
  \begin{tabular}{|c|l|l|l|l|}
    \hline
    Constraints & most (freq.) & median & mean & std dev. \\
    \hline
    AllDifferent & 0 ~~~~~~~~~(98)& 0 & 5.001 & 35.185 \\
    LinearSum & 0.004 ~~(48)& 0.059 & 0.032 & ~~0.027 \\
    Minimum & 0 ~~~~~~~~~(71)& 0 & 0.026 & ~~0.044 \\
    NoOverlap & 0.039 ~~(32)& 0.074 & 0.074 & ~~0.030\\
    Ordered & 0.020 (100)& 0.020 & 0.020 & ~~0\\
    \hline
  \end{tabular}
  \caption{Training errors (100 runs)  of Experiment~1, over small and
    complete constraint assignment spaces.}
  \label{tab:training}
\end{table}

\begin{table}
  \centering
  \small
  \begin{tabular}{|c|c|l|l|l|l|}
    \hline
    Exp. & Constraints & most freq & median & mean & std dev \\
    \hline
    \multirow{5}{*}{1} &
    AllDifferent & 0 & 0 & 0.017 & 0.119 \\
    &LinearSum & 3$\times10^{-4}$ & 0.019 & 0.179 & 0.341 \\
    &Minimum & 0 & 0 & 1.435 & 4.866 \\
    &NoOverlap & 0.268 & 0.316 & 0.486 & 0.682\\
    &Ordered & 0.106 & 0.106 & 0.106 & 0\\
    \hline
    \multirow{5}{*}{2} &
    AllDifferent & 0.052 & 0.052 & 0.052 & 0 \\
    &LinearSum & 3$\times10^{-4}$ & 3$\times10^{-4}$ & 0.200 & 0.629 \\
    &Minimum & 0 & 0 & 0.193 & 0.978 \\
    &NoOverlap & 0.202 & 0.202 & 0.215 & 0.020\\
    &Ordered & 0.050 & 0.050 & 0.054 & 0.008\\
    \hline
  \end{tabular}
  \caption{Test errors  in high dimensions of  error functions learned
    with Experiments~1 and~2.}
  \label{tab:test}
\end{table}

\subsubsection{Experiment 3}

The goal of this experiment is  not to be state-of-the-art in terms of
run-times for solving Sudoku, but  to compare the average run-times of
the same  solver on four  nearly identical Sudoku models  presented in
Section~\ref{sec:xp_proto3}. For  the model with a  hand-crafted error
function, we implemented the \emph{primal graph based violation error}
of  AllDifferent   from  Petit  et   al.~\shortcite{Petit2001}.   This
function simply  outputs the number  of couples with  identical values
within  a given  assignment.   To  run this  experiment,  we used  the
framework \ghost from Richoux et al.~\shortcite{GHOST}, which includes
a local search algorithm able to handle both \csp and \efsp models.

\begin{table}
  \centering
  \small
  \begin{tabular}{|l|c|r|r|r|}
    \hline
    Sudoku & Error Function & mean & median & std dev \\
    \hline
    \multirow{4}{*}{$9\times9$} &
    nothing (\csp)& \cellcolor{gray!15}624.41 & \cellcolor{gray!15}217.21 & \cellcolor{gray!15}1,1196.80 \\
    &hand-crafted    & 33.86  & 32.00  & 10.34 \\
    &fast-forward    & 55.57  & 49.27  & 41.17 \\
    &hard-coded      & 33.48  & 31.76  & 9.48 \\
    \hline
    \multirow{4}{*}{$16\times16$} &
    nothing (\csp)& \cellcolor{gray!15}- & \cellcolor{gray!15}- & \cellcolor{gray!15}- \\
    &hand-crafted & 432.34 & 393.95 & 164.06 \\
    &fast-forward    & 825.61  & 774.36  & 271.09 \\
    &hard-coded      & 537.48 & 539.86 & 162.03 \\
    \hline
  \end{tabular}
  \caption{Mean  run-times  in milliseconds  over  100  runs to  solve
    Sudoku  with  4  different  representations  of  the  AllDifferent
    constraint. Rows  in gray means  that some runs hit  the 10-second
    timeout.}
  \label{tab:sudoku}
\end{table}

Table~\ref{tab:sudoku} shows that models  with error functions clearly
outperformed  the   model  with   the  constraint  represented   as  a
predicate. Over 100  runs, no error function-based models  hit the 10s
timeout, but 4 runs of the  regular constraint network model timed out
on  the \(9\times9\)  grid,  and  all of  them  on the  \(16\times16\)
grid. Moreover, the learned error function hard-coded in C++ is nearly
as  efficient  as the  hand-crafted  one  (also  coded in  C++).   The
difference of  runtimes between the learned  error function hard-coded
and computed  through the  \icn gives  us an idea  of the  overhead of
computing such a function through the \icn.

\section{Discussions}\label{sec:discussions}




Like Freuder~\shortcite{Freuder2007} wrote: ``\textit{This research program
  is not easy because 'ease of  use' is not a science}.''  However, we
believe our  result is a step  toward the 'ease of  use' of Constraint
Programming, and in particular about \efsp and \efop. 
Our  system  is  excellent  for learning  error  functions  of  simple
constraints  over  complete  spaces. For  intrinsically  combinatorial
constraints,  learning   over  large,  incomplete  spaces   should  be
favored. 
One of the  most significant results in this paper  is that our system
outputs  interpretable  results,   unlike  regular  artificial  neural
networks.  Error functions output by our system are intelligible. This
allows our  system to have two  operating modes: 
\begin{inparaenum}[1.]
  \item a  fully automatic system,  where error functions  are learned
    and  called within  our  system, being  completely transparent  to
    users  who  only need  to  furnish  a  concept function  for  each
    constraint, in addition  to the regular sets of  variables \(V\) and
    domains \(D\), and
  \item a  decision support system, where  users can look at  a set of
    proposed error functions, pick up and modify the one they prefer.
\end{inparaenum}   


The current  limitation of our  system is  that it struggles  to learn
high-quality error  function for very combinatorial  constraints, such
as Ordered and, in particular, NoOverlap1D.  By combining results from
Experiments~1  and~2, we  can conclude  that:  1.  our  system is  not
overfitting but need more diverse and expressive operations to learn a
high-quality error function  for such constraints, and  2. the Hamming
cost is certainly  not the better choice to  represent their assignment
error.

An extension of our work would  be to do reinforcement learning rather
than supervision learning based on  the Hamming cost.  
Learning  via   reinforcement  learning  would  allow   finding  error
functions that are  more adapted to the chosen solver.

\bibliographystyle{named}
\bibliography{learning_ef}

\begin{thebibliography}{}

\bibitem[\protect\citeauthoryear{Beldiceanu and Simonis}{2012}]{Beldiceanu12}
Nicolas Beldiceanu and Helmut Simonis.
\newblock A model seeker: Extracting global constraint models from positive
  examples.
\newblock In {\em Principles and Practice of Constraint Programming ({CP}
  2012)}, pages 141--157. Springer, 2012.

\bibitem[\protect\citeauthoryear{Beldiceanu \bgroup \em et al.\egroup
  }{2007}]{catalog}
Nicolas Beldiceanu, Mats Carlsson, Sophie Demassey, and Thierry Petit.
\newblock Global constraint catalog: Past, present and future.
\newblock {\em Constraints}, 12:21--62, 2007.

\bibitem[\protect\citeauthoryear{Bessiere \bgroup \em et al.\egroup
  }{2005}]{Bessiere05}
Christian Bessiere, Remi Coletta, Fr{\'{e}}d{\'{e}}ric Koriche, and Barry
  O'Sullivan.
\newblock A sat-based version space algorithm for acquiring constraint
  satisfaction problems.
\newblock In {\em 16th European Conference on Machine Learning ({ECML} 2005)},
  pages 23--34. Springer, 2005.

\bibitem[\protect\citeauthoryear{Bessiere \bgroup \em et al.\egroup
  }{2007}]{Bessiere07}
Christian Bessiere, Remi Coletta, Barry O'Sullivan, and Mathias Paulin.
\newblock Query-driven constraint acquisition.
\newblock In {\em Proceedings of the 20th International Joint Conference on
  Artificial Intelligence ({IJCAI} 2007)}, pages 50--55. {IJCAI/AAAI} Press,
  2007.

\bibitem[\protect\citeauthoryear{Bessiere \bgroup \em et al.\egroup
  }{2013}]{Bessiere13}
Christian Bessiere, Remi Coletta, Emmanuel Hebrard, George Katsirelos, Nadjib
  Lazaar, Nina Narodytska, Claude{-}Guy Quimper, and Toby Walsh.
\newblock Constraint acquisition via partial queries.
\newblock In {\em Proceedings of the 23rd International Joint Conference on
  Artificial Intelligence ({IJCAI} 2013)}, pages 475--481. {IJCAI/AAAI} Press,
  2013.

\bibitem[\protect\citeauthoryear{Bessiere \bgroup \em et al.\egroup
  }{2017}]{Bessiere2017}
Christian Bessiere, Frederic Koriche, Nadjib Lazaar, and Barry O'Sullivan.
\newblock Constraint acquisition.
\newblock {\em Artificial Intelligence}, 244:315--342, 2017.

\bibitem[\protect\citeauthoryear{Boussemart \bgroup \em et al.\egroup
  }{2016}]{XCSP-paper}
Frederic Boussemart, Christophe Lecoutre, Gilles Audemard, and Cédric Piette.
\newblock {XCSP3: An Integrated Format for Benchmarking Combinatorial
  Constrained Problems}.
\newblock {\em {arXiv e-prints}}, abs/1611.03398:1--238, 2016.

\bibitem[\protect\citeauthoryear{Caniou \bgroup \em et al.\egroup
  }{2015}]{caniou2015constraints}
Yves Caniou, Philippe Codognet, Florian Richoux, Daniel Diaz, and Salvador
  Abreu.
\newblock Large-scale parallelism for constraint-based local search: The costas
  array case study.
\newblock {\em Constraints}, 20(1):30--56, 2015.

\bibitem[\protect\citeauthoryear{Codognet and Diaz}{2001}]{AS}
Philippe Codognet and Daniel Diaz.
\newblock Yet another local search method for constraint solving.
\newblock In {\em International Symposium on Stochastic Algorithms: Foundations
  and Applications ({SAGA} 2001)}, pages 73--90. Springer, 2001.

\bibitem[\protect\citeauthoryear{Cooper \bgroup \em et al.\egroup
  }{2020}]{GuidedTour}
Martin Cooper, Simon Givry, and Thomas Schiex.
\newblock Valued constraint satisfaction problems.
\newblock In {\em A Guided Tour of Artificial Intelligence Research}, volume~2,
  pages 185--207. Springer, 2020.

\bibitem[\protect\citeauthoryear{Freuder}{1997}]{Freuder1997}
Eugene~C. Freuder.
\newblock In pursuit of the holy grail.
\newblock {\em Constraints}, 2(1):57--61, 1997.

\bibitem[\protect\citeauthoryear{Freuder}{2007}]{Freuder2007}
Eugene~C. Freuder.
\newblock Holy grail redux.
\newblock {\em Constraint Programming Letters}, 1:3--5, 2007.

\bibitem[\protect\citeauthoryear{Freuder}{2018}]{Freuder2018}
Eugene~C. Freuder.
\newblock Progress towards the holy grail.
\newblock {\em Constraints}, 23(2):158--171, 2018.

\bibitem[\protect\citeauthoryear{Frisch \bgroup \em et al.\egroup
  }{2008}]{essence}
Alan Frisch, Warwick Harvey, Chris Jefferson, Bernadette Martínez-Hernández,
  and Ian Miguel.
\newblock {ESSENCE}: A constraint language for specifying combinatorial
  problems.
\newblock {\em Constraints}, 13:268--306, 2008.

\bibitem[\protect\citeauthoryear{Hurley \bgroup \em et al.\egroup
  }{2016}]{Hurley16}
Barry Hurley, Barry O'sullivan, David Allouche, George Katsirelos, Thomas
  Schiex, Matthias Zytnicki, and Simon~De Givry.
\newblock Multi-language evaluation of exact solvers in graphical model
  discrete optimization.
\newblock {\em Constraints}, 21(3):413–434, 2016.

\bibitem[\protect\citeauthoryear{Hutter \bgroup \em et al.\egroup
  }{2009}]{paramils}
Frank Hutter, Holger~H. Hoos, Kevin Leyton-Brown, and Thomas St\"{u}tzle.
\newblock {ParamILS:} an automatic algorithm configuration framework.
\newblock {\em Journal of Artificial Intelligence Research}, 36:267--306, 2009.

\bibitem[\protect\citeauthoryear{Keijzer \bgroup \em et al.\egroup }{2002}]{EO}
Maarten Keijzer, J.~J. Merelo, G.~Romero, and M.~Schoenauer.
\newblock {Evolving Objects: A General Purpose Evolutionary Computation
  Library}.
\newblock {\em Artificial Evolution}, 2310:829--888, 2002.

\bibitem[\protect\citeauthoryear{Nethercote \bgroup \em et al.\egroup
  }{2007}]{minizinc}
Nicholas Nethercote, Peter~J. Stuckey, Ralph Becket, Sebastian Brand,
  Gregory~J. Duck, and Guido Tack.
\newblock Minizinc: Towards a standard cp modelling language.
\newblock In {\em Principles and Practice of Constraint Programming ({CP}
  2007)}, pages 529--543. Springer Berlin Heidelberg, 2007.

\bibitem[\protect\citeauthoryear{Petit \bgroup \em et al.\egroup
  }{2001}]{Petit2001}
Thierry Petit, Jean-Charles Régin, and Christian Bessiere.
\newblock Specific filtering algorithms for over-constrained problems.
\newblock In {\em International Conference on Principles and Practice of
  Constraint Programming (CP 2001)}. Springer, 2001.

\bibitem[\protect\citeauthoryear{Puget}{2004}]{Puget2004}
Jean-François Puget.
\newblock Constraint programming next challenge: Simplicity of use.
\newblock In {\em {International Conference on Principles and Practice of
  Constraint Programming (CP 2004)}}, pages 5--8. Springer, 2004.

\bibitem[\protect\citeauthoryear{Richoux \bgroup \em et al.\egroup
  }{2016}]{GHOST}
Florian Richoux, Alberto Uriarte, and Jean-François Baffier.
\newblock {GHOST}: A combinatorial optimization framework for real-time
  problems.
\newblock {\em IEEE Transactions on Computational Intelligence and AI in
  Games}, 8(4):377--388, 2016.

\bibitem[\protect\citeauthoryear{Stanley}{2007}]{CPPN}
Kenneth~O. Stanley.
\newblock {Compositional Pattern Producing Networks: A Novel Abstraction of
  Development}.
\newblock {\em {Genetic Programming and Evolvable Machines}}, 8(2):131--162,
  2007.

\bibitem[\protect\citeauthoryear{Teso}{2019}]{Teso19}
Stefano Teso.
\newblock Constraint learning: An appetizer.
\newblock In {\em Reasoning Web: Explainable Artificial Intelligence}, pages
  232--249. Springer, 2019.

\bibitem[\protect\citeauthoryear{Wallace}{2003}]{Wallace2003}
Mark Wallace.
\newblock Languages versus packages for constraint problem solving.
\newblock In {\em {International Conference on Principles and Practice of
  Constraint Programming (CP 2003)}}, pages 37--52. Springer, 2003.

\end{thebibliography}

\end{document}